# BMI: A Behavior Measurement Indicator for Fuel Poverty Using Aggregated Load Readings from Smart Meters

Paul Fergus and Carl Chalmers

Department of Computer Science, Liverpool John Moores University, Byrom Street, L3 3AF, P.Fergus, C.Chalmers {@ljmu.ac.uk}

**Abstract.** Fuel poverty affects between 50 and 125 million households in Europe and is a significant issue for both developed and developing countries globally. This means that fuel poor residents are unable to adequately warm their home and run the necessary energy services needed for lighting, cooking, hot water, and electrical appliances. The problem is complex but is typically caused by three factors; low income, high energy costs, and energy inefficient homes. In the United Kingdom (UK), 4 million families are currently living in fuel poverty. Those in series financial difficulty are either forced to self-disconnect or have their services terminated by energy providers. Fuel poverty contributed to 10,000 reported deaths in England in the winter of 2016-2107 due to homes being cold. While it is recognized by governments as a social, public health and environmental policy issue, the European Union (EU) has failed to provide a common definition of fuel poverty or a conventional set of indicators to measure it. This chapter discusses current fuel poverty strategies across the EU and proposes a new and foundational behavior measurement indicator designed to directly assess and monitor fuel poverty risks in households using smart meters, Consumer Access Device (CAD) data and machine learning. By detecting Activities of Daily Living (ADLS) through household appliance usage, it is possible to spot the early signs of financial difficulty and identify when support packages are required.

**Keywords.** Energy efficient homes, Energy Tariffs, Fuel Poverty, Policy, Measurement, Load Monitoring.

## 1.1 Introduction

Fuel poverty describes members of a household that cannot afford to adequately warm their home or run the necessary energy services needed for lighting, cooking, hot water, and electrical appliances [1]. It is estimated that between 50 and 125 million households are affected in Europe (EPEE, 2009). In the UK, approximately 4 million households are classified as being fuel poor (15% of all households) – 613,000 in Scotland (24.9% of the total); 291,000 in Wales (23% of the total); 160,000 in Northern Ireland (22% of the total); and 2.55 million in England (11% of the total) [2]. The problem is complex but is typically caused by three factors; low income, high energy costs, and energy inefficient homes [1], [3]–[5].

In the UK, financial support is provided for low-income households through the Warm Home Discount Scheme, Cold Weather Payments and Winter Fuel Payments (similar support is provided in other EU member states) [6]. According to a UK report written in 2018, the government provided £1.8 billion in funding annually for Winter Fuel Payments, £320 million for the Warm Homes Discount Scheme, and £600 million for the Energy Company Obligation scheme [7]. Schemes like this provide temporary relief, but do not tackle the underlying causes of fuel poverty [8], [9].

Currently, fuel bills in the UK cost on average £1,813 a year, a 40% increase from £1,289 in 2015 [10]. The Office of Gas and Electricity Markets (Ofgem) caps the maximum price that consumers can pay for electricity and gas, however, the recent lifting of price caps has seen a £1.7bn increase in consumer bills [11]. Subsequently, rising energy prices force more people to live in fuel poverty rather than easing the financial pressures fuel poor households already have [12].

Alongside low income and rising fuel costs, a substantial share of the residential housing stock in Europe is older than 50 years with many in use reportedly hundreds of years old [13]. More than 40% were constructed before the 1960's when energy regulations were limited [14]. The performance of buildings depends on the installed heating system and building envelope, climatic conditions, indoor temperatures and fuel poverty [15]. This means that largest energy savings often come from improving older buildings, particularly poorly insulated properties built before the 1960s.

In the UK, the energy efficiency of homes is measured using the Standard Assessment Procedure (SAP) rating [16]. During the winter months colder weather lowers the energy efficiency of the property and increases domestic energy demand. The performance of the heating system, appliances, and the number of people living in the property (and how long they say in the home throughout the day) determine the household fuel bill. In low-income and energy inefficient homes the winter months are particularly problematic and
a source of constant worry for occupants about debt, affordability, and ther-
mal discomfort [17]. The impact this has on health is significant given that
fuel poor households spend increased amounts of time in the cold. Hence,
poor health among this social group is prevalent [18]. In fact, evidence
shows us that fuel poor occupants are more likely to experience poor health;
miss school [19]–[24]; and report absences from work [25], [17].

According to the E3G, the UK has the sixth-highest rate of Excessive
Winter Deaths (EWD) of the 28 EU member states - a large number have
been directly linked to cold homes [19], [26]. EWD is the surplus number
of deaths that occur during the winter season (in the UK this is between the
22nd of December and 20th of March) compared with the average number
of deaths in non-winter seasons [19]. The main causes of EWD are circula-
tory and respiratory diseases [27]. It is estimated that about 40% of EWD
are attributable to cardiovascular diseases, and 33% to respiratory diseases
[22]. According to the Office of National Statistics (ONS), there were
50,100 EWDs in England and Wales in the 2017-2018 winter period; the
highest recorded since the winter of 1975-1976 [28]. Cold homes have also
been linked with high blood pressure [29], heart attacks and pneumonia,
particularly among vulnerable groups such as children and older people
[22], [23], [30]–[33]. This often leads to inhabitants experiencing loss of
sleep, increased stress and mental illness [17].

Alongside serious health outcomes, cold homes are uninviting leaving in-
habitants stigmatized, isolated, and embarrassed because they are often
forced to put on additional clothing, wrap up in duvets or blankets and use
hot water bottles to stay warm [34]. This undoubtably increases the likeli-
hood of depressions and other mental illness. Epidemiological studies show
that occupants in damp homes are more likely to have poorer physical and
mental health [35]. According to the Building Research Establishment
(BRE) poor housing costs the National Health Service (NHS) £1.4 billion
each year [36]. The World Health Organization (WHO) commissioned a
comprehensive analysis of epidemiological studies and concluded that a re-
lationship exists between humidity and mold in homes and health-related
problems [37].

Fuel poverty is a focal point for the EU however as the figures show,
current policy has had/is having little effect on reducing the number of fuel
poor households. This is hardly surprising given the EU does not provide a
common definition of fuel poverty or a set of indicators to measure it [38].
This means that fuel poverty numbers vary depending on what measurement
indicator is implemented.

## 1.2  Measuring Fuel Poverty

Measurement indicators are used to identify which households are considered to be in fuel poverty – in the UK, this is the responsibility of the Department for Business, Energy & Industrial Strategy (BEIS) [39]. A detailed report, commissioned by the EU in 2014, found that 178 indicators exist: of which 58 relate to income or expenditure and 51 to physical infrastructure [40]. Indicators related to energy demand and demographics amount to 10 and 15 respectively. 139 are single metric indicators and 39 combinatory or constructed indicators, representing 22% of the total and mostly falling under the category of income/expenditure. Among the identified energy poverty metrics, 10 are consensual-based; 42 expenditure-based and 11 outcome-based; while another 14 indicators comprise a combination of metrics. The two main approaches used today are expenditure-/consensual-based. Only the most common indicators within both approaches will be considered in this chapter. For a more detailed discussion the reader is referred to [40].

### 1.2.1 Expenditure-Based Indicators

Expenditure-based indicators focus primarily on the proportion of the household budget used to pay for domestic fuel [41]. The best-known indicator is the 10% rule proposed by Boardman in the early 1990s [1] which was adopted in the UK in 2001. A household is classed as being fuel poor if more than 10% of its income is spent on fuel to maintain an acceptable heating regime [42]. The indicator uses a ratio of modelled fuel costs and a Before Housing Costs (BHC) measure of income [43]. Modelled fuel costs are derived from energy prices and a modelled consumption figure that includes data about property size, the number of people in the property, the household's energy efficiency rating and the types of fuel used. Fuel poor households are those with a ratio greater than 1:10 (10%).

The Hills report in 2011, commissioned by the Department of Energy & Climate Change (DECC) (now BEIS), triggered a replacement of the 10% indicator with the Low Income High Cost (LIHC) indicator [44]. LIHC is now used in the UK to measure fuel poverty and has attracted considerable attention within different national contexts [43], [45], [46], [47]. The LIHC indicator is calculated using a national income threshold and a fuel cost threshold [42], [44]. A household is classified as fuel poor if it exceeds both thresholds. The fuel cost threshold is a weighted median of the fuel costs for all households, weighted according to the number of people in a property. This average fuel cost value is the assumed cost of achieving an adequate level of comfort. The threshold is the same for all households of equivalent size. The income threshold is calculated as 60% of the weighted national median for income After Housing Costs (AHC) are accounted for. The income figure for each household is also weighted to account for the number of people living in the property. This figure is combined with the weighted fuel costs of the household. The income threshold is therefore higher for those that require a greater level of income to meet larger fuel bills.

### 1.2.2 Consensual-Based Indicators

Consensual-based indicators on the other hand assess whether a person is in fuel poverty by asking them. The approach was initially based on Townsend's early relative poverty metric [48] and later on the consensual poverty indicator proposed in [49] and [50]. The fundamental principle is centered on a person's inability '*to afford items that the majority of the general public considered to be basic necessities of life*' [50].

Using surveys, household occupants are asked to make subjective assessments about their ability to maintain and adequately warm their home and pay their utility bills on time. The EU has adopted the core principles of the consensual model and implemented the Survey on Income and Living Conditions (EU-SILC) [51]. EU-SILC includes a set of questions that asks whether the household is able to a) keep their home warm during winter days, b) has been in arrears with utility bills, and c) whether the house has leakages or damp walls [52]. The recommendation was launched in 2003 and was the first micro-level data set to provide data on income and other social and economic aspects of people living in the EU [51].

EU-SILC has a rotating panel that lasts four years; a quarter of the sample is replaced each year by new subsample members [53]. During the four years households are contacted up to four times. The consensual approach has been acclaimed for being easy to implement and less complex, in terms of collecting data, than expenditure-based indicators. A key feature of the EU-SILC dataset is that it provides an important basis for identifying and understanding fuel poverty and the differences that exist across all EU member states [54].

### 1.2.3 Limitations

Fuel poverty measures have several limitations, primarily because of the multi-dimensional nature of the phenomenon, which makes it difficult to adequately capture or measure it using a single indicator [40]. Additionally, most indicators have been disparaged for focusing solely on fuel expenditure without consideration for under-consumption which has led to governments underestimating the real extent of fuel poverty [55], [44]. In the case of expenditure-based approaches, the main issue is the lack of available data, particularly on the contributing factors needed to assess the extent of fuel poverty. This is alleviated with consensus-based approaches given that micro-level data is collected. However, the approach has also been criticized for being too subjective and exclusive [56].

In the case of the 10% rule, it does not respond to variations in income, fuel prices or energy efficiency improvements [57] and this has led to skewed results [58]. Hills suggested that '*flaws in the 10% indicator have distorted policy choices, and misrepresented the problem*'. Therefore, relatively well-off households in energy inefficient properties have been identified as being fuel poor [59], [57].

The LIHC indicator on the other hand excludes low-income, single person households [59], [60]. Moore argues that this indicator obscures increases in energy prices, as its introduction has led to a fall in fuel poor households, in spite of significant increases in energy costs during the same period [58]. This has been described by some as an attempt to move the goalposts in order to justify missing targets for the eradication of fuel poverty, which was a target for all households by 2016 [61]. Middlemiss, adds that the LIHC priorities energy efficiency as a solution to fuel poverty distracting from other drivers, such as the wider failure of the energy market to provide an affordable, and appropriate energy supply to homes [62].

Finally the EU-SILC consensus-based approach has been criticized for a) only including specific household types, b) containing anomalies in the data collected (i.e. missing data), c) being subjective due to self-reporting, and d) providing a limited understanding of the intensity of the issue due to the binary character of the metrics [56]. Participants do not view judgements like 'adequacy of warmth; in the same way while some households may not even identify themselves as being fuel poor due to pride even though they have been characterized as being fuel poor under other measures [56]. It is not unusual for fuel poor residents to deny the reality of their situation, and report that they are warm enough when they are in fact not.

## 1.3 Smart Meters

Residential homes consume 23% of the total energy delivered worldwide (29% in the UK) [63]. Industries consume 37% and this is closely followed by transportation which is 28% [64]. Household energy consumption is considered a multidimensional phenomenon rooted within a socio-cultural and infrastructure context, and as such occupant behavior is complex. Existing measurement indicators, as we have seen, fail to capture the behavioral traits associated with individual households. Yet, with the current smart meter rollout well underway in many developed countries which facilitates the automatic reporting of energy usage, it is now possible to capture the behavioral aspects of energy consumption through data provided by CADs paired with smart meters [65]. CADs provide data every 10 seconds for all energy consumed within the home at the aggregated level [66]. This data combined with advanced data analytics allows us to determine whether a house is occupied, what electrical appliances are operated, and when they are being used [67],[68]. Such insights provide the based for routine formation which we will return to later in the chapter.

### 1.3.2 Smart Meter Infrastructure

Smart meters measure gas and electricity consumption and send usage information to energy suppliers and other interested parties. This, a) removes the need for home visits and manual meter readings and b) allows consumption data to be used by the smart grid, to balance energy load and improve efficiency [69]. According to the International Energy Agency (IEA), smart grids are essential to meet future energy requirements [70], given that worldwide energy demand is expected to increase annually by 2.2%, eventually doubling by 2040 [71].

Energy consumption data in the smart grid is received directly from smart meters and stored, managed and analyzed in the Meter Data Management System (MDMS) [66]. The MDMS is implemented in the data and communications layer of the Advanced Metering Infrastructure (AMI) and is a scalable software platform that provides data analytic services for AMI applications, i.e., data and outage management, demand and response, remote connect/disconnect, smart meter events, and billing [66]. Data contained in the MDMS is shared with consumers, market operators and regulators.

Smart meters in the UK collect and transmit energy usage data to the MDMS every 30 minutes [72]. Higher sample rates are possible, but this increases the costs for data storage and processing. Data transmitted through a smart meter consists of a) aggregated energy data in watts (W), b) a Unix date/time stamp and c) the meters personal identification number (PID). The energy distribution and automation system, collects data from sensors dispersed in the smart grid. Each sensor generates up to 30 readings per second and includes a) voltage and equipment health monitoring, and b) outage voltage and reactive power management information. External data sets by third party providers are also used to facilitate demand and response subsystems. OS/firmware software provides a communication link between the MDMS and smart technologies and this allows geographically aggregated load readings to be analysed to ensure efficient grid management. The OS/firmware system also manages OS/firmware version patching and updating. Figure 1 shows a typical MDMS system and its common components.

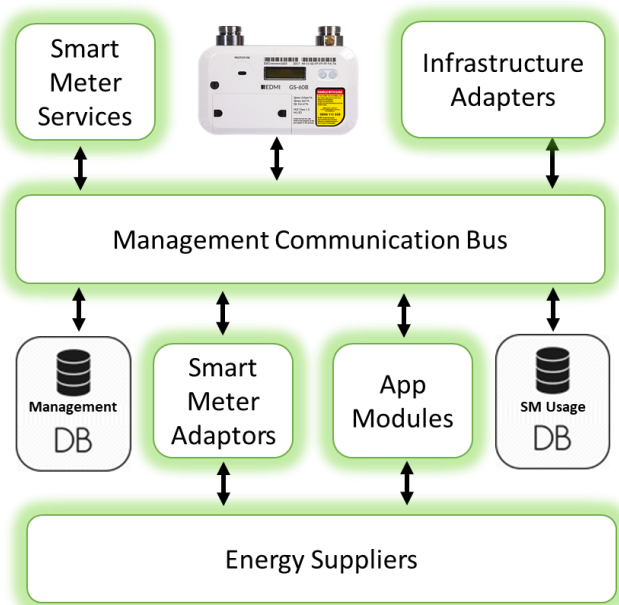

**Fig. 1:** Meter Data Management System for Processing Home Energy Usage and Automated Billing

Information stored in the MDMS is a significant data challenge that requires data science tools to maintain optimal operational function [73], [74] and derive insights from the information collected [75], [76]. This allows decision making and service provisioning to be implemented directly atop the smart meter infrastructure [77]–[81]. Services exploit the smart grid infrastructure to provide application support in different domains, i.e. health, climate change, and energy optimization [82].

### 1.3.3 Smart Meter Sampling Frequencies

Most studies do not use actual smart meter data for monitoring. Smart meter readings are provided every 30-minutes in the UK (other countries have different sample frequencies) [83]. With 30-minute data it is possible to detect occupancy; however no reliable appliance information can be noticed at this frequency [84]. Therefore, electricity monitors are either paired with the smart meter using a consumer access device (CAD), CT Clip, or sensor plugs attached to the actual appliance when higher sample frequencies are required as shown in figure 2.

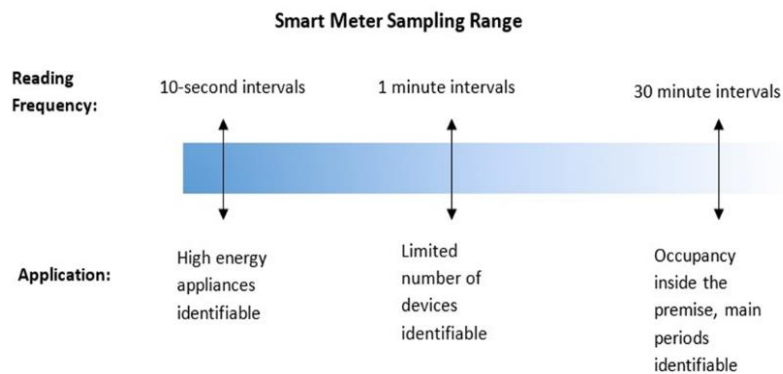

**Fig. 2:** Capabilities Based on Sampling Frequency

CADs are an inexpensive way to obtain whole-house measurements at higher sampling rates (i.e. readings every 10 seconds in the UK). With a CAD you can detect when high energy appliances, such as an oven, kettle and microwave, are being operated. CT Clips are used when either a smart meter has not yet been installed in a household or when sample frequencies higher than every 10 seconds are required. CT Clips, clamped around the power cable (live or natural), can sample the aggregated energy feed thousands of times every second. Though, the approach is more costly than a CAD as additional hardware and software need to be installed. With a CT Clip, it is possible to detect faulty appliances and overlapping use, including low energy appliances, such as lights and audio equipment. Device types will be discussed in more detail later in the chapter.

### 1.3.4 Load Disaggregation

Load disaggregation is a broad term used to describe a range of techniques for splitting a household's energy supply into individual electrical appliance signatures, for example, a kettle, microwave or oven [68]. There are a number of reasons why load disaggregation is important. In the context of fuel poverty, appliance detections provide the basis for habitual appliance usage patterns, which manifest as routine household behaviours [68], [83]. Through an understanding of normal routine behaviour it is possible to identify anomalies and assess whether they are linked to fuel poverty indictors – more on this later [83].

Disaggregating electrical device usage is called Appliance Load Monitoring (ALM) [85]. ALM is divided into two types: Non-Intrusive Load Monitoring (NILM) [86] and Intrusive Load Monitoring (ILM) [87]. NILM is a single point sensor, such as a smart meter or CT clip. In contrast, ILM is a distributed sensing method that uses multiple sensors – one for each electrical device being monitored [87]. ILM is more accurate than NILM as energy usage is read directly from sensors attached to each electrical appliance being measured. The practical disadvantages however include high costs, multi sensor configuration and complex installation [88]. More importantly, ILM sensors can be moved between different devices and this can skew identification and classification results.

NILM on the other hand is less accurate than ILM and more challenging as appliances are identified from aggregated household energy readings [89]. NILM was first developed in the mid 80's [90]. Since then academic interest in the field has increased rapidly [91]. More recently there has been significant commercial interest [92]. This has been primarily driven by an increased focus on energy demand combined with significant reductions in the cost of sensing technology, and equally, improvements in machine learning algorithms. Commercial interest is directly linked with the huge commercial potential of services that exploit the smart metering infrastructure, for example in health, energy management, and climate change.

### 1.3.5 Electrical Device Types

Electrical appliances, alongside their normal on-off states, run in multiple modes. Many devices have low power requirements or standby modes, while appliances like ovens operate using several control functions. Understanding different device categories is important in NILM, as they define different electrical usage characteristics. Device categories include, Type 1,

Type 2, Type 3, and Type 4. The associated signals for each are illustrated
in Figure 3.

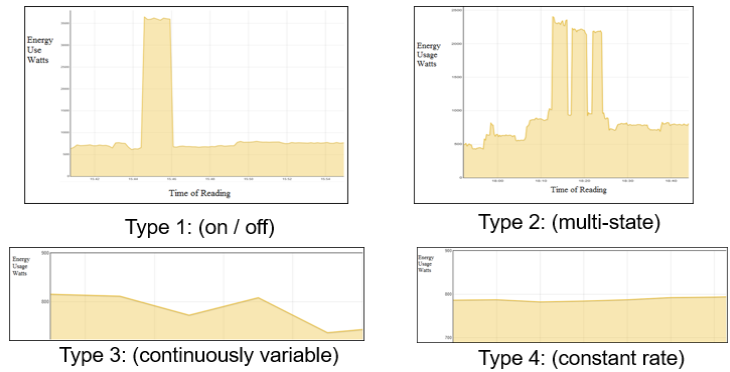

Type 1: (on / off)    Type 2: (multi-state)

Type 3: (continuously variable)    Type 4: (constant rate)

**Fig. 3:** Appliance Type Energy Readings

The characteristics for each appliance type are described as:

- Type 1 devices are either on or off. Examples include kettles, toast-
  ers and lighting. Figure 4 illustrates a power reading for a kettle –
  (a) shows a series of devices being used in conjunction or in close
  succession; while (b) presents evenly distributed single device in-
  teractions.
- Type 2 devices, known as Multi-State Devices (MSD) or finite state
  appliances, operate in multiple states and have more complex be-
  haviours than Type 1 devices. Devices include washing machines,
  dryers and dishwashers.
- Type 3 devices, known as Continuously Variable Devices (CVD),
  have no fixed state. There is no repeatability in their characteristics
  and as such they are problematic in NILM. Example devices include
  power tools such as a drill or electric saw.
- Type 4 are fairly new in terms of device category. These devices are
  active for long periods and consume electricity at a constant rate –
  they are always on. Hence, there is no major events to detect other
  than small fluctuations. Such devices include smoke detectors and
  intruder alarms.

Understanding device types is important in any load disaggregation sys-
tem, as electrical appliances are often used in combination, typically when
preparing meals. This can affect the performance in classification tasks due
to the boundaries that exist between device classes, making them difficult to identify. The boundaries between classes provide guidance on what classifiers to use (i.e. linear, quadratic or polynomial) within the same feature space [93].

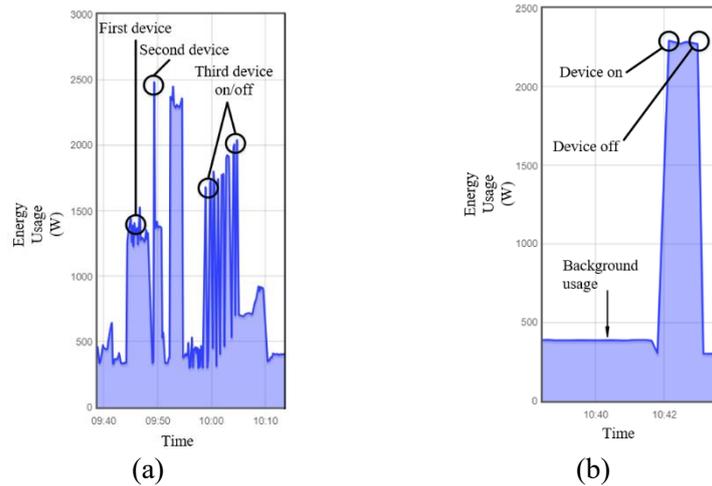

(a)                                          (b)

**Fig. 4:** Aggregated Load Readings Highlighting Unique Device Signatures

## 1.4  BMI: A Behavior Measurement Indicator for Fuel Poverty Assessments

Measuring and monitoring household fuel poverty is challenging as we have seen [40]. Expenditure-based approaches lack data on all the contributing factors needed to sufficiently assess the extent of fuel poverty. Using this method, the data is often derived from a subjective and generalized view of households, including their occupants and how energy is consumed. In fact, data is often skewed or contaminated given that households may not even identify themselves as being in fuel poverty due to pride [94]. The remainder of this chapter proposes a different point of view that incorporates personalized household behavior monitoring using activities of daily living. By doing this it is possible to understand the unique characteristics of each household in terms of what, when and how often electrical appliances are used. The hope is to derive some useful insights and provide a more objective measure of fuel poverty from a socio- behavioral view point to better support the occupants and their energy needs.

### 1.5.1 BMI Framework

The Behavior Measurement Indicator (BMI) proposed was initially developed and evaluated in partnership with Mersey Care NHS Foundation Trust to measure appliance usage in dementia patients and derive routine behaviors for social care support [95], [83]. Here we consider an extension to the existing framework and build on the behavioral monitoring aspects of the system to provide a household BMI indicator for fuel poverty assessment.

The BMI builds on the existing smart meter infrastructure. Smart meters in households, paired with a CAD using the ZigBee Smart Energy Profile (SEP) [96], provide access to aggregated power usage readings every 10 seconds. This sample frequency allows high powered appliances associated with ADLs to be detected and used to establish household behavioural routines. Appliances, such as a kettle, microwave, washing machine and oven are regarded as necessary appliances used by occupants to live a normal life (ADLs). Therefore, appliances such as TVs, mobile chargers, computers, and lighting are of limited interest as they do not contribute to ADL assessment, for example, TVs are often left on for background noise and provide no information about what an occupant in a household is doing [83].

The BMI operates in three specific modes in order to achieve this; device training mode; behavioural training mode; and prediction model.

- **In device training mode** power readings are obtained from the CAD and recorded to a data store. Readings, alongside device usage annotations are used to train the machine learning algorithms to classify appliances from aggregated load readings. Features automatically extracted using a one-dimensional convolutional neural network (discussed in more detail later in the chapter) act as input vectors to a fully connected multi-layer perceptron (MLP) for device classification.

- **In behavioural training mode** features from device classifications are extracted to identify normal and abnormal patterns in behaviour. The features allow the system to recognise the daily routines performed by occupants in a household, including their particular habits and behavioural trends.

- **In prediction mode** both normal and abnormal household behaviours are detected and remediated.

The framework implements web services for machine-to-machine communications using enterprise ready protocols, Application Programming Interfaces (API's) and standards. The monitoring application interfaces with web services to receive real-time monitoring alerts about the household's status (i.e. green for normal behaviour, amber for unusual behaviour, and red when drastic changes occur). The complete end-to-end system is shown in Figure 5.

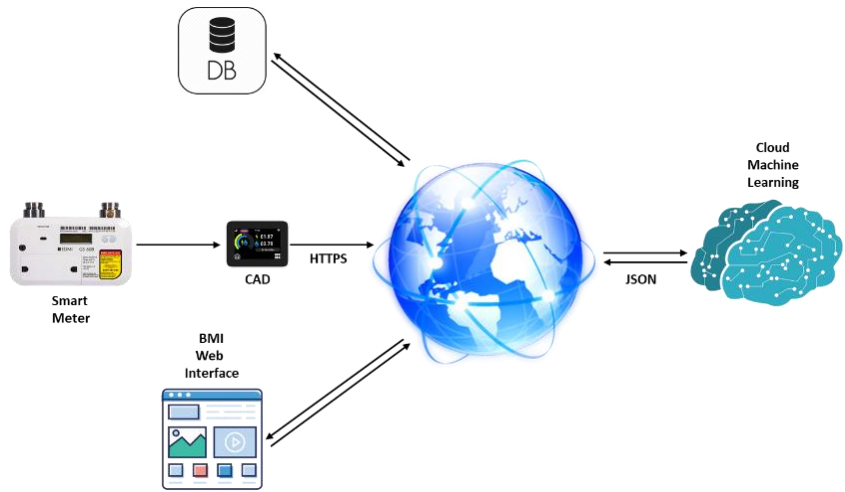

**Fig. 5:** System Framework Showing the end-to-end components

### 1.5.2 Data Collection

The training dataset for device classification is constructed using energy monitors (i.e. a CAD paired with a household smart meter). CAD payload data contains the aggregated energy readings generated every 10 seconds. To detect ADLs, a kettle, microwave, washing machine, oven and toaster are used, although others could be included if required, such as an electric shower depending on the relapse indicators of interest in fuel poverty.

Generating device signatures is achieved using a mobile app to record when each appliance is operated (annotation). Time-stamped recordings are compared with mobile app recordings to extract specific appliance signatures. Each signature is labelled and added to the training data and subsequently used to train the machine learning algorithms for appliance classification.

### 1.5.3 Data Pre-Processing

CAD energy readings are filtered and transformed before they are used to train machine learning algorithms. A high-pass filter is implemented to remove background noise below 300 watts (although this value needs to be
personalised based on individual household energy usage as each home will
be different) – signals below this threshold typically represent Type 4 elec-
trical appliances which cannot be detected using CAD data.

Device signatures are obtained by switching appliances on and off indi-
vidually and filtering normal background noise. Individual appliance signa-
tures are combined to generate new appliance usage patterns that represent
composite appliance usage. For example, Figure 6 shows that when the in-
dividual energy readings for three appliances (kettle, microwave and toaster)
are combined (i.e. they are operated in parallel) a 'Total Load' signature is
produced.

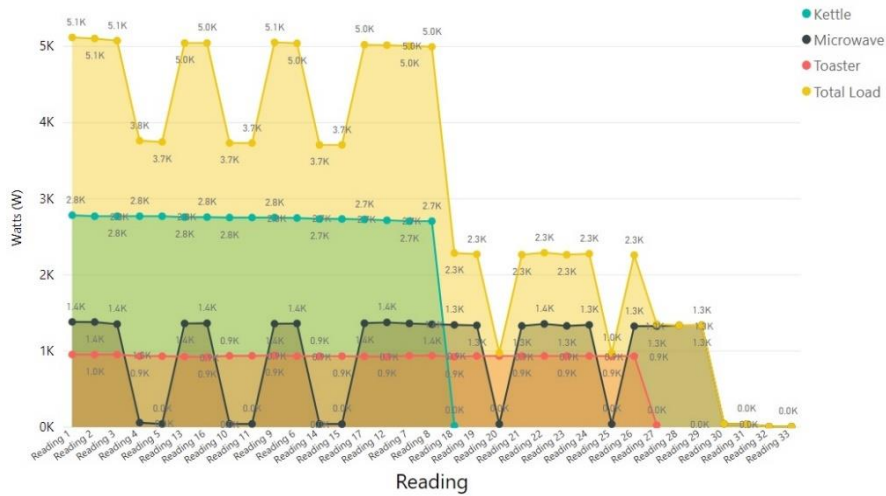

**Fig. 6:** Whole household aggregated power consumption and individual
device power consumption

The aggregated signature (total load) describes the three appliances being
used in parallel. Repeating this process for all device combinations yields
different aggregate signatures that describe which devices are on and which
are not. Hence, a dataset is built containing individual and combined appli-
ance usage signatures and used train and detect which of the ADL appliances
are in use.

### 1.5.4 CAD NILM Machine Learning Model for Appliance Disaggregation

In contrast to manually extracted features based on input from domain knowledge experts, (i.e. peak frequency and sample entropy) features can automatically learn from appliance energy signatures using a one dimensional convolutional neural network (1DCNN) [97]. Appliance signatures are input directly to a convolutional layer in the 1DCNN. The convolutional layer detects local features along the time-series signal and maps them to feature maps using learnable kernel filters (features). Local connectivity and weight sharing are adopted to minimise network parameters and overfitting [98]. Pooling layers are implemented to reduce computational complexity and enable hierarchical data representations [98]. A single convolutional and pooling layer pair along with a fully connected MLP comprising two dense layers and softmax classifier output (an output for each appliance being classified) completes the 1DCNN network as the time-signals are not overly complex. The proposed architecture is represented in Figure 7.

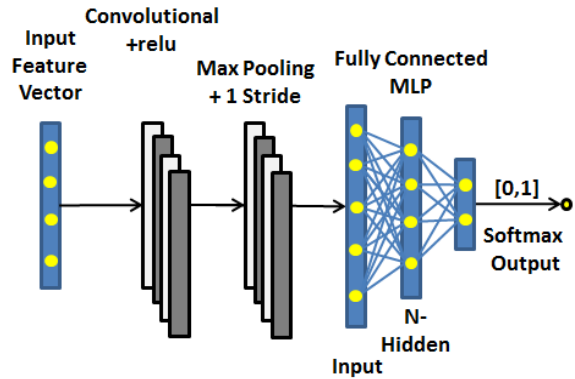

**Fig. 7:** One Dimensional Convolutional Neural Network

The network model is trained by minimizing the cost function using feedforward and backpropagation passes. The feedforward pass constructs a feature map from the previous layer to the next through the current layer until an output is obtained. The input and kernel filters of the previous layer are computed as follows:

$$z_j^l \sum_{l-1}^{M^{l-1}} 1dconv\left(x_i^{l-1}, k_{ij}^{l-1}\right) + b_j^l$$

Where $x_j^{l-1}$ and $Z_j^l$ are the input and output of the convolutional layer, respectively, and $k_{ij}^{l-1}$ the weight kernel filter from the $i^{th}$ neuron in layer $l$ -1 to the $j^{th}$ neuron in layer $l$, $1dconv$ represents the convolutional operation and $b_j^l$ describes the bias of the $j^{th}$ neuron in layer $l$. $M^{l-1}$ defines the number of kernel filters in layer $l-1$. A ReLU activation function is used for transforming the summed weights and is defined as:

$$x_j^l = ReLU(z_j^l)$$

Where $x_j^l$ is the intermediate output at current layer $l$ before downsampling occurs. The output from current layer $l$ is defined as:

$$y_j^l = downsampling\left(x_j^l\right) \quad x_j^{l+1} = y_j^l$$

Where *downsampling()* represents a max pooling function that reduces the number of parameters, and $y_j^l$ is the output from layer $l$ and the input to the next layer $l$ +1. The output from the last pooling layer is flattened and used as the input to a fully connected MLP. Figure 8 shows the overall process.

The error coefficient $E$ is calculated using the predicted output $y$:

$$E = -\sum_n \sum_i (Y_{ni} \log(y_{ni}))$$

Where $Y_{ni}$ and $y_{ni}$ are the target labels and the predicted outputs, and $i$ the number of classes in the $n^{th}$ training set. The learning process optimizes the network free parameters and minimises $E$. The derivatives of the free parameters are obtained and the weights and biases are updated using the learning rate ($\eta$). To prompt rapid convergence, *Adam* is implemented as an optimisation algorithm and *He* for weight initialisation. The weights and bias in the convolutional layer and fully connected MLP layers are updated using:

$$k_{ij}^l = k_{ij}^l - \eta \frac{\partial E}{\partial k_{ij}^l} b_j^l = b_j^l - \eta \frac{\partial E}{\partial b_j^l}$$

Small learning rates reduce the number of oscillations and allow lower error rates to be generated. Rate annealing and rate decay are implemented to address the local minima problem and control the learning rate change across all layers.

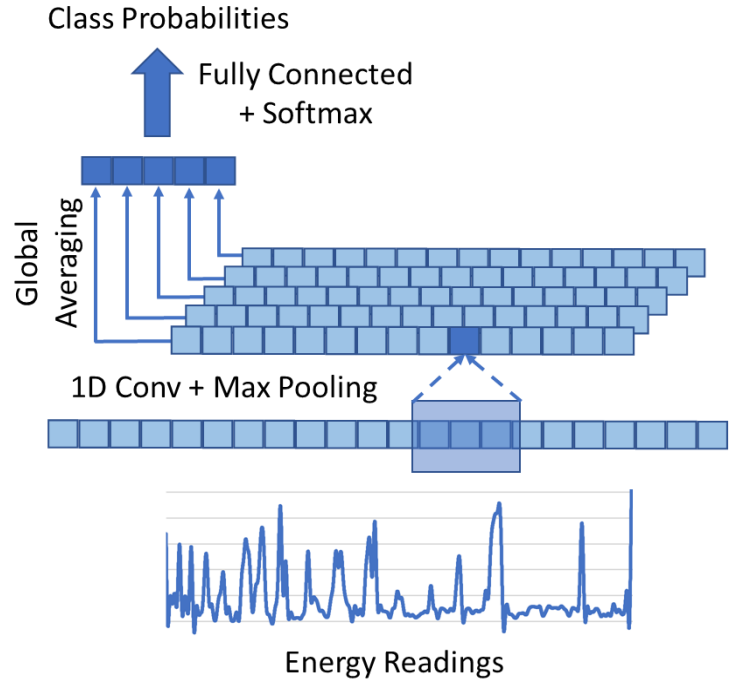

**Fig. 8:** Convolution and Max Pooling Process

Momentum start and ramp coefficients are used to control momentum when training starts and the amount of learning for which momentum increases - momentum stable controls the final momentum value reached after momentum ramp training examples. Complexity is controlled with an optimised weight decay parameter, which ensures that a local optimum is found.

The number of neurons and hidden layers required to minimise $E$, including activation functions and optimisers, can be determined empirically. Input and hidden layers are also determined empirically depending on data and the number of softmax outputs required for classification. The network free parameters can be obtained using the training and validation sets over a set number of epochs and evaluated with a separate test set comprising unseen data.

The 1DCNN approach allows the unique features from single appliance and composite appliance energy signatures to be automatically extracted and used in subsequent machine learning modelling for classification tasks. This removes the need for manual feature engineering and simplifies the data analysis pipeline.

### 1.5.5 Measuring Behaviour

Current fuel poverty measurement indicators cannot directly collect, monitor or assess fuel poverty in households in real-time. ADL is a term used in healthcare to assess a person's self-care activities. [99]. With smart meters, CADs and 1DCNNs, the BMI platform can analyse electrical appliance interactions and detect ADLs (routine behaviours) in all households connected to the smart grid using smart meters [78]–[80], [84]. Household occupants carry out ADLs each day as part of their normal routine behaviour. These include preparing breakfast, lunch and dinner, making cups of tea, switching on lights, and having a shower. While such tasks are common to us all there will be differences. For example, one household may use the toaster to make toast for breakfast; while another might use the cooker to make porridge. Some might boil the kettle to make tea in the evening after finishing work, others might prefer to have a glass of wine. Some households might use the shower (likely at different times of the day and frequency, i.e. one or two showers a day), others might prefer to have a bath.

These activities can be easily detected through ongoing interactions with home appliances. This is useful for deriving normal routine behaviours within households, but more importantly to detect anomalies, for the purpose of safeguarding vulnerable homes against fuel poverty risks. How we interact and use energy in our home will likely be affected by our circumstances, i.e. having a baby, children moving out of the family home, gaining employment (or losing a job) as well as caring for an elderly family member who has moved in.

Such circumstantial changes directly alter our routine use of electrical appliances. For example, in the case of having a baby, the microwave, kettle or oven hob may be operated throughout the night for a period of time to heat the milk required to bottle feed babies. In the unfortunate situation where a person has lost their job, household occupants may have to substitute fresh food cooked using the oven and hob for more cheaper food options, such as microwave meals. These are clues, that household circumstances have changed. Families experiencing financial difficulties may have to cut heating-based appliance usage and ration hot water – this will lead to an overall dip in energy consumed by that household.

Significant changes in behaviour will act as key indicators and facilitate decision-making strategies to support struggling households. For example, appliances operated during abnormal times of the day (when this is not normal behaviour for that household) may indicate that occupants are experiencing difficulties (i.e. making tea in the early hours of the morning could be due to sleep disturbances possibly caused through financial worry;

conversely occupants staying in bed for longer periods of time or not cooking meals may indicate severe financial difficulty or energy disconnection issues). The BMI system can detect significant changes in behaviour like these as we see in the next section.

### 1.5.5.1 Vectors for Behavioural Analysis

Individual device detections classified by the CAD NILM machine learning model are combined as feature vectors for behaviour analysis. Predicted classes are given a unique device ID and assigned to an observation window depending on the time of day the appliance is used, i.e. during breakfast or eventing meal times.

Observation windows can be defined and adjusted to meet the unique behaviours of each household. This is performed automatically following a baseline learning period for each household connected to the smart grid. Observation windows capture routine behaviour and act as placeholders for the fuel poverty relapse indicators being measured and monitored (these will need to be defined by fuel poverty experts). This allows the system to construct a personalised representation of each household and assign device usage to specific observation windows common to that household. Continually repeating this process allows routine behaviours to be identified and anomalies in behaviour to be detected. Figure 9 describes 7 possible observation windows in a 24-hour period. Each observation window is configurable to meet the unique needs of the application or service.

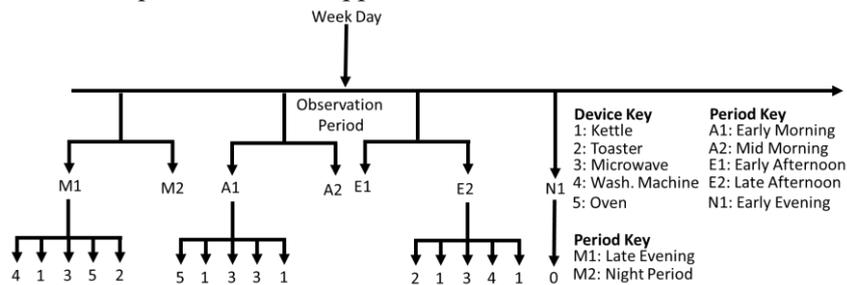

**Fig. 9:** Device Assignment for Identifying Key Activities within Significant Observation Periods

The order of device interactions is not necessarily important unless there is a clear deviation from normal behaviour. From the behaviour vectors it is possible to see the degree of correlation between appliance usage and the hour-of-day (strong routine behaviour). Figure 10 shows the correlations for different home appliances used over a 6 month period [100].

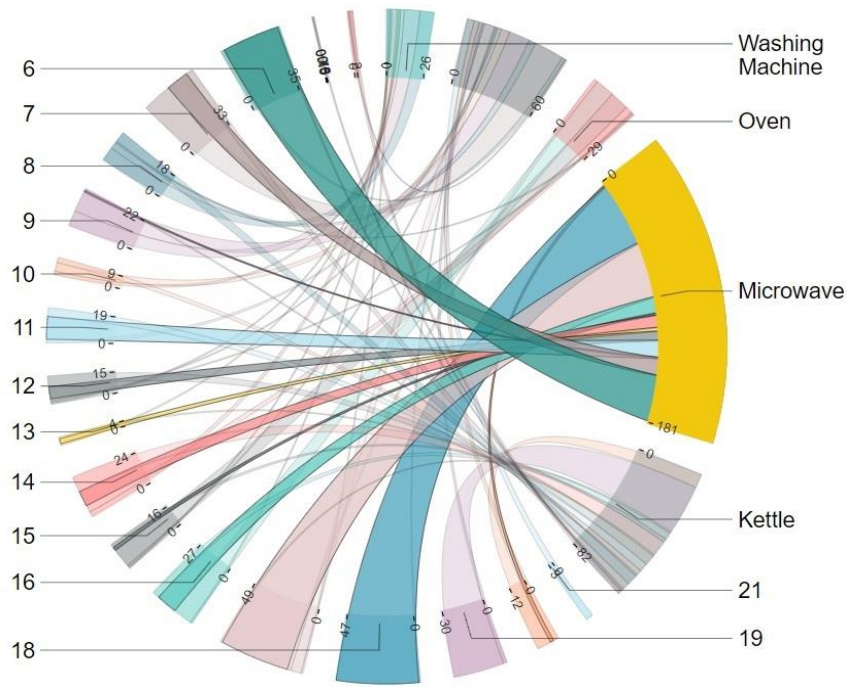

**Fig. 10:** Degree of Correlation between Device Usage and Hour

The figure shows quantitative information relating to flows, including re-
lationships and transformations. The lines between appliances and time-of-
day, like ant pheromone trails, show the established routine behaviour for a
particular home. For example, it is possible to see that the microwave is
mostly used at 06:00 hours and 18:00 hours. Alternations in either link pro-
portionality or association may indicate the early signs of circumstantial
change which could be linked to fuel poverty risk factors. Anomalies are
progressed through a traffic light system - red would suggest a sustained
change in routine behaviour over a period of time (time period would be set
by expert in fuel poverty) and may or may not indicate that the house is
experiencing financial difficulties. Conversely, green would show that nor-
mal routine behaviour has been observed and that no support or intervention
is required. Amber would flag the house as worrisome (this does not neces-
sarily mean the house is transitioning into a fuel poverty state, simply that a
change in behaviour has been detected). This could be caused by circum-
stantial changes, i.e. people coming to stay or household occupants going
on holiday. Viewing Figure 6 periodically we would expect to see changes between correlations and their associated strengths for those households experiencing significant changes in normal routine behaviours.

Anomalies in device usage can be seen with the Z-score technique to describe data points in terms of their relationship to the mean and the standard deviation of a group of points. Figure 7 shows the inliers in green which represent normal appliance interactions for that household. Each cluster represents a specific appliance class. The outliers are depicted in red where both the kettle and toaster classes in this case reside outside the household's normal routine behaviour. Figure 11 shows that in total three kettles were used on three separate occasions between the hours of 00:00 and 05:00 and a single interaction with a toaster was detected during the same observation period. In the context of fuel poverty such results may provide interesting insights when managing fuel poverty households. As the household continues to struggle financially, we would expect routine behaviour to become more erratic (or even disappear for long periods) leading to an increase in the number of anomalies detected.

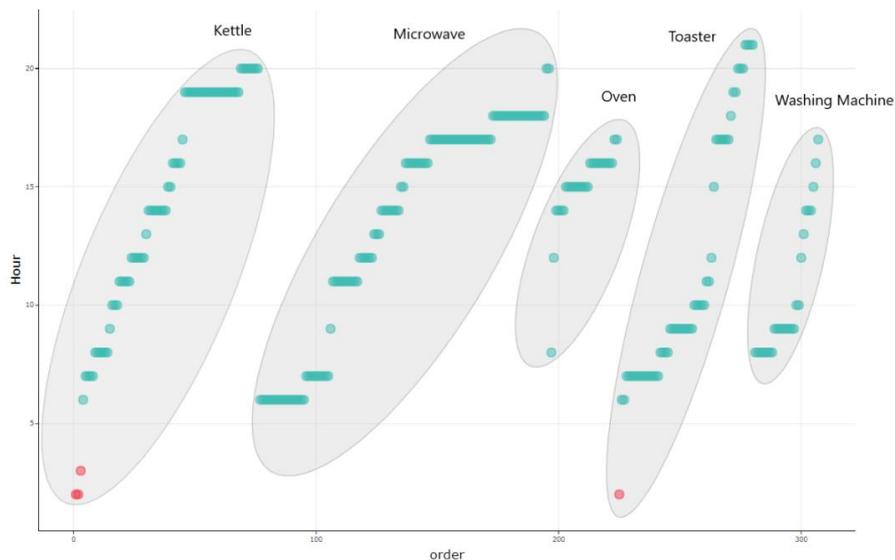

**Fig. 11:** Sleep Disturbances for an Occupant using Z-Score Anomaly Detection

The BMI framework presents the first platform of its kind that capitalises on the smart meter infrastructure to describe a behaviour measurement indicator for use in fuel poverty assessments. It has been designed to exploit the smart metering infrastructure and provide foundational services to more accurately assess fuel poverty in real-time within individual households [77]. Obviously, future trials are required to test the applicability of the BMI system and evaluate whether it has any real potential in tackling fuel poverty. Based on our previous use of the system in dementia, the technology is a powerful tool for assessing routing behaviour and detection anomalies. We therefore think the solution will lend itself to household behaviour analysis (in terms of electricity consumption) in fuel poverty assessment [83].

The use of association rule mining within load disaggregation is also interesting technique that can uncover relationships and their associated strengths using transactional data. Identifying device relationships (what devices are commonly used together or in sequence) and their relationship with the time of day can expose strong behavioural traits within the dwelling. Reoccurring deviation from identified routine patterns or the weakening of common relationships could be used to trigger an intervention where fuel poverty is suspected. Figure 12 highlights the use of association rule mining to determine the relationship strength between an appliance and time of day.

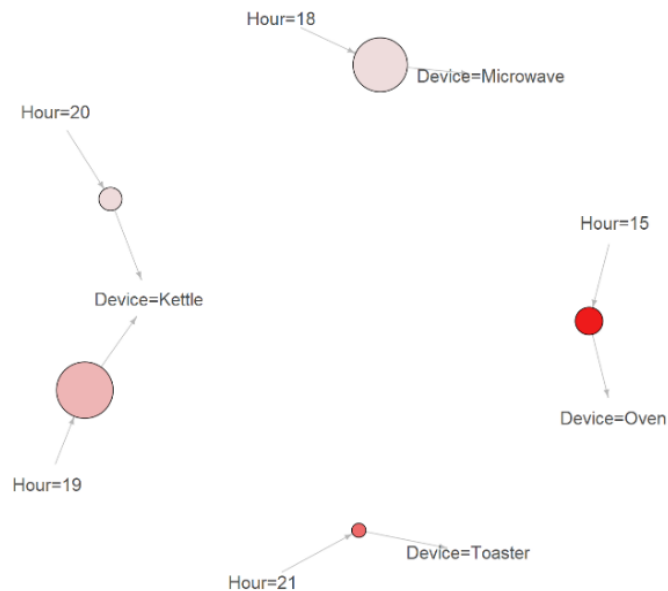

**Fig. 12:** Association Rule Mining for the Identification of Behavioural Patterns

Association rule mining can be used to provide a more abstracted view above and beyond the aggregated load level of a dwelling. Instead, the collective behaviour of entire regions could be monitored to assess the impact of shifting financial and social economic changes. For example, raising fuel prices or the closure of large employers (retail/manufactures) and reduction in the associated foot flow to a region. By using association rule mining the impact can be objectively measured and the effectiveness of any intervention/recovery passively monitored.

## 1.6 Discussion

As this chapter has highlighted fuel poverty affects a significant number of households in Europe and indeed globally. The problem is primarily caused by a combination of low income, high energy costs, and energy inefficient homes. In the UK, 4 million households are currently in fuel poverty, which, among other things, contributes to poor health and premature winter deaths. Poor quality housing has also been linked with fuel poverty which is hardly surprising given that a substantial share of the residential stock in Europe is older than 50 years.

The problem is recognized by governments; however, the EU has not yet adopted a common definition of fuel poverty, nor a set of common indicators to measure it, making a standardised approach difficult to implement. Many households move in and out of fuel poverty but there are households that find themselves persistently trapped in fuel poverty [101]. Measuring and monitoring fuel poverty is challenging as we have seen [40] and while Expenditure-based approaches have been proposed they lack data on all the contributing factors needed to sufficiently assess fuel poverty. Consensus-based approaches on the other hand have data, but this is only from snap shots in time meaning data is often outdated, subjective and exclusive in nature.

Of the 178 measurement indicators reported in the literature, many do not respond to variations in income, circumstantial changes, fuel prices or energy efficiency improvements. They exclude low-income and single person households [59], [60] and this has distorted policy choices, and misrepresented the problem. Against this negative backdrop and an overall distrust of government bodies and energy providers, fuel poor customers feel that the intensity of the issue is not fully understood by those developing policies to combat it [56].

We proposed the BMI system to monitor a household's activities of daily living and understand routine behaviour in order to gain insights into how energy is consumed [78]–[80], [84]. Households behave in different ways. While there may be common tasks, such as meal preparation, there will be differences. By detecting ADLs using appliance interactions, it is possible to derive routine behaviour for each household. This makes BMI highly personalised and sensitive to the unique characteristics of each household connected to the smart grid.

Changes in behaviour can be identified and investigated and support services provided if and when they are needed. Modelling ADLs in households will allow the onset of fuel poverty issues to be identified much earlier. When households are identified, appropriate packages can be put in place to help mitigate the adverse effects fuel poverty has on fuel poor occupants. Detecting self-disconnect in households, particularly among the most vulnerable in society, such as young children and the elderly, would allow appropriate support services to be put in place to ensure homes are appropriately warm.

The identification of expected behaviour and relapse indicators aids in the selection of appropriate analytical techniques. Establishing routines facilitates the detection of abnormal behaviour. Combining this with unique energy signatures within each household a new and foundational fuel poverty indicator is possible that is adaptable and reflective of household circumstances. We believe that the BMI system could contribute significantly to the fuel poverty domain. To the best of our knowledge BMI is the first of its kind as currently there is no fuel poverty measurement indicator that can measure household energy usage interactions and derive routine behaviour in every home fitted with a smart meter. The approach is highly personalised and closely aligned with the different routines' households exhibit despite the size of the house or the number of occupants. Once routine behaviour has been established, BMI is highly sensitive to change; using a traffic light system it is therefore possible to target and support households classified as being fuel poor.

## 1.7 Conclusions

This chapter discussed the many aspects of fuel poverty and the government policies put in place to combat it. The key message is that cold homes waste energy and harm their occupants. Most fuel poor indicators are derived from generalised estimates disconnected from the unique characteristics of individual households. Houses and occupants do not behave the same – they have their own socio-behavioural characteristics that affect how and when energy is consumed. Therefore, coupled with the household envelope and the many other factors that influence household behaviours, there is a disparity between existing measurement indicators and fuel poverty prevalence.

The only way to fully understand fuel poverty is to measure high-risk households and the unique characteristics and behaviours they exhibit in terms of energy consumption and ADLs. We believe that the BMI system can do this will minimal installation requirements as the solution exploits the existing smart meter infrastructure to provide appropriate services. System operation requires no input from household occupants as BMI is based on the assessment of ADLs (the everyday things that people do in their home in order to survive) captured through normal appliance interactions.

The BMI has been previously evaluated in a clinical trial with Mersey Care NHS Foundation Trust to model the ADLs of dementia patients [83]. However, it has been possible to extend the system to include fuel poverty risk factors following minor changes to observation periods and fuel poverty related relapse indicators. Future work will focus on a trial to evaluate the BMI system in fuel and non-fuel poverty homes. Cases will include households that find themselves in and out of fuel poverty. Controls will be those households that have not previously experienced fuel poverty or had difficulties with paying bills and keeping their home warm. The measurable outputs will be to evaluate whether the BMI system can detect which houses are in or likely to be in fuel poverty and those that are not.

To the best of our knowledge this is the first fuel poverty measurement indicator that builds on the existing smart meter infrastructure and associated CAD technology to carry out NILM and personalised ADL monitoring in every household connected to the smart grid that is designed to safeguard households and occupants against fuel poverty.

***Acknowledgement.*** This work was inspired by an event run in Liverpool where the authors were invited to present at the "Better at Home" workshop run by the National Energy Action (NEA) organization. We would like to particularly thank Matt Copeland and Dr Jamie-Leigh Rosenburgh at NEA who asked whether we could extend our dementia smart meter framework to include a behavior measurement indicator for fuel poverty.